\documentclass{bmvc2k}

\usepackage{graphicx, wrapfig, lipsum}
\DeclareMathOperator*{\argmin}{argmin} 

\title{Mutual Suppression Network for Video Prediction using Disentangled Features}

\addauthor{Jungbeom Lee}{jbeom.lee93@snu.ac.kr}{1}
\addauthor{Jangho Lee}{ubuntu@snu.ac.kr}{1}
\addauthor{Sungmin Lee}{simon0810@snu.ac.kr}{1}
\addauthor{Sungroh Yoon $^*$}{sryoon@snu.ac.kr}{1}

\addinstitution{
 $^1$ Department of Electrical and \\Computer Engineering\\
 Seoul National University\\
 Seoul 08826, Korea
}
\addinstitution{
\vspace{7pt}
 $^*$ Corresponding Author
}

\runninghead{Lee et al.}{Video Prediction using Disentangled Features}

\begin{document}

\maketitle

\begin{abstract}
Video prediction has been considered a difficult problem because the video contains not only high-dimensional spatial information but also complex temporal information. Video prediction can be performed by finding features in recent frames, and using them to generate approximations to upcoming frames. We approach this problem by disentangling spatial and temporal features in videos. We introduce a mutual suppression network (MSnet) which are trained in an adversarial manner and then produces spatial features which are free of motion information, and motion features with no spatial information. MSnet then uses motion-guided connection within an encoder-decoder-based architecture to transform spatial features from a previous frame to the time of an upcoming frame. We show how MSnet can be used for video prediction using disentangled representations. We also carry out experiments to assess the effectiveness of our method to disentangle features. MSnet obtains better results than other recent video prediction methods even though it has simpler encoders.
\end{abstract}

\vspace{-10pt}
\section{Introduction}
Given a sequence of frames from a video, the process of video prediction attempts to generate one or more upcoming frames.
Video prediction is important in real-time systems such as robots, closed-circuit television (CCTV), and self-driving cars, and also has a place in applications such as the unsupervised learning of image representations from videos~\cite{srivastava2015unsupervised}.

The learning of representations from images has been studied extensively, and the results now surpass human ability~\cite{he2015delving}. However, learning representations from videos remains a challenging task because of the temporal dimension, which brings a huge number of variations, and because it is not possible to annotate every frame in a video with labels.
Some `natural' labeling of videos is possible, for instance based on temporal coherence. However, the entangling of content and motion information in videos tends to make unsupervised learning challenging.
In this regard, there have been previous works on decomposing videos into content and motion components~\cite{simonyan2014two,tulyakov2017mocogan,villegas2017decomposing,wang2016temporal,wang2016actions,feichtenhofer2016spatiotemporal}. While the learning techniques used on images~\cite{he2016deep,huang2017densely} can be extended for the content representations~\cite{misra2016shuffle,wang2015unsupervised}, the learning of representations of motion has not been studied so extensively. Temporal information can be obtained from optical flow \cite{ng2015beyond}, with reasonable results. However, optical flow estimation involves a great deal of computation and depends on having a labeled dataset, which requires tremendous effort and cost to obtain.

We propose a technique in which a mutual suppression network (MSnet) is used to disentangle motion and content features. This approach is based on the following intuitive assertions:

\noindent\textbf{Separability of features: }We train MSnet in such a way that information of one type is suppressed during the extraction of features of another type. This can be achieved by mutual adversarial learning.

\noindent\textbf{Content from several frames: }The majority of methods that encode video into motion and content obtain content features from a single frame. We argue that content features, as well as motion features, should be obtained from several frames. A single frame is not sufficient to capture content information if two objects are occluded or cannot be distinguished.

\noindent\textbf{Reproducibility: }Given three frames $x^{(1)}$, $x^{(2)}$, and $x^{(3)}$, the content features from $x^{(1)}$ and $x^{(2)}$, together with the motion features from $x^{(2)}$ and $x^{(3)}$, should allow us to reproduce $x^{(3)}$. This leads to motion and content features which contain semantic information.

\noindent\textbf{Time-reversibility of content: }While previous methods have been based on the assumption that content features are mostly time-invariant, we propose that the content features should be time-reversible, so a content feature obtained from $(x_1, x_2)$ should be the same as that obtained from $(x_2, x_1)$. This time-reversible property is intended to ensure that motion information is not unwittingly included in content features because we extract content features from two frames, which may be related by temporal information.

The second step is the frame prediction task using the encoders and a generator trained in the first step. To generate frames from the features from the encoders, previous methods utilize the skip connections as used in UNet~\cite{ronneberger2015u}, that transfers information direct from a previous frame to a target frame. During frame prediction, however, it is better that the generator takes information related to the target frame, not the previous frame. Therefore we introduce a motion-guided connection which modifies the information from the previous frame to become the information needed for generating the target frame by considering the motion features.

\vspace*{-10pt}
\section{Related Work}
There is no easy way of representing spatial and temporal information simultaneously in videos.
Recent work in video representation learning has therefore focused on disentangling temporal and spatial information in natural videos. 
Simonyan and Zisserman used a two-stream network for action recognition in videos, motivated by the way in which the human visual cortex decouples complementary information appearing in videos~\cite{simonyan2014two}, which has subsequently been used for various fields of video processing~\cite{jain2017fusionseg, yang2017spatio, chung2017two}. 

The prediction of video frames requires the ability both to understand previous frames and to produce realistic new frames. These tasks can be facilitated by decomposing a video into motion and content components using techniques based on a two-stream network. VGAN~\cite{vondrick2016generating} predicted upcoming frames by modeling the foreground separately from the background. MCnet \cite{villegas2017decomposing} used an encoder-decoder technique to separate the motion and content information of a video: a content encoder extracts spatial features from the most recent frame of a video, and a motion encoder captures motion dynamics from pixel-wise differences between previous pairs of frames. However, few of these differences contain any semantic information about motions. 
DRnet~\cite{denton2017unsupervised} used a content discriminator to separate the pose attributes from the content attributes in a frame. The content discriminator examines whether two pose features relate to the same content or not. The pose features acquired in this way are used to predict future pose features, from which upcoming frames can be generated. However, DRnet cannot catch pure content information because it only uses one-way suppression. For example, when predicting the frames using videos in the KTH dataset~\cite{schuldt2004recognizing}, DRnet sometimes changed the identities of human in the predicted frames. In addition, poses tend to be more ambiguous than motions in videos, so DRnet sometimes swapped the locations of two numerals when applied to the Moving MNIST dataset~\cite{srivastava2015unsupervised}. DRnet is only concerned with a series of absolute locations (poses), and not with relative locations (motion).

\vspace*{-10pt}

\section{Proposed Method}


The proposed method consists of two steps. The first step is frame reproduction, which obtains disentangled features from a consideration of semantics in the frames. The second step is video prediction using the disentangled features obtained during frame reproduction.


\vspace*{-10pt}
\subsection{Frame Reproduction} 
\label{featureextractors}
Let $x_t$ denote the $t^{\text{th}}$ frame in video $x$. The frame reproduction is to reproduce $x_{t+k}$ from the known frames $x_t, x_{t+1}$, and $x_{t+k}$. Following the `Reproducibility' assertion presented in the Introduction, we reproduce $x_{t+k}$ from the content of $x_t, x_{t+1}$ and the motion of $x_{t+1}, x_{t+k}$, with the aim of obtaining disentangled motion and content features by considering semantics. We describe our network architecture in Section~\ref{networkarchi} and our training procedure in Section~\ref{trainingprocedure}.
\vspace*{-10pt}
\subsubsection{Network Architecture}\label{networkarchi}
\begin{wrapfigure}{r}{5.5cm}
  \centering
  \includegraphics[width=0.88\linewidth]{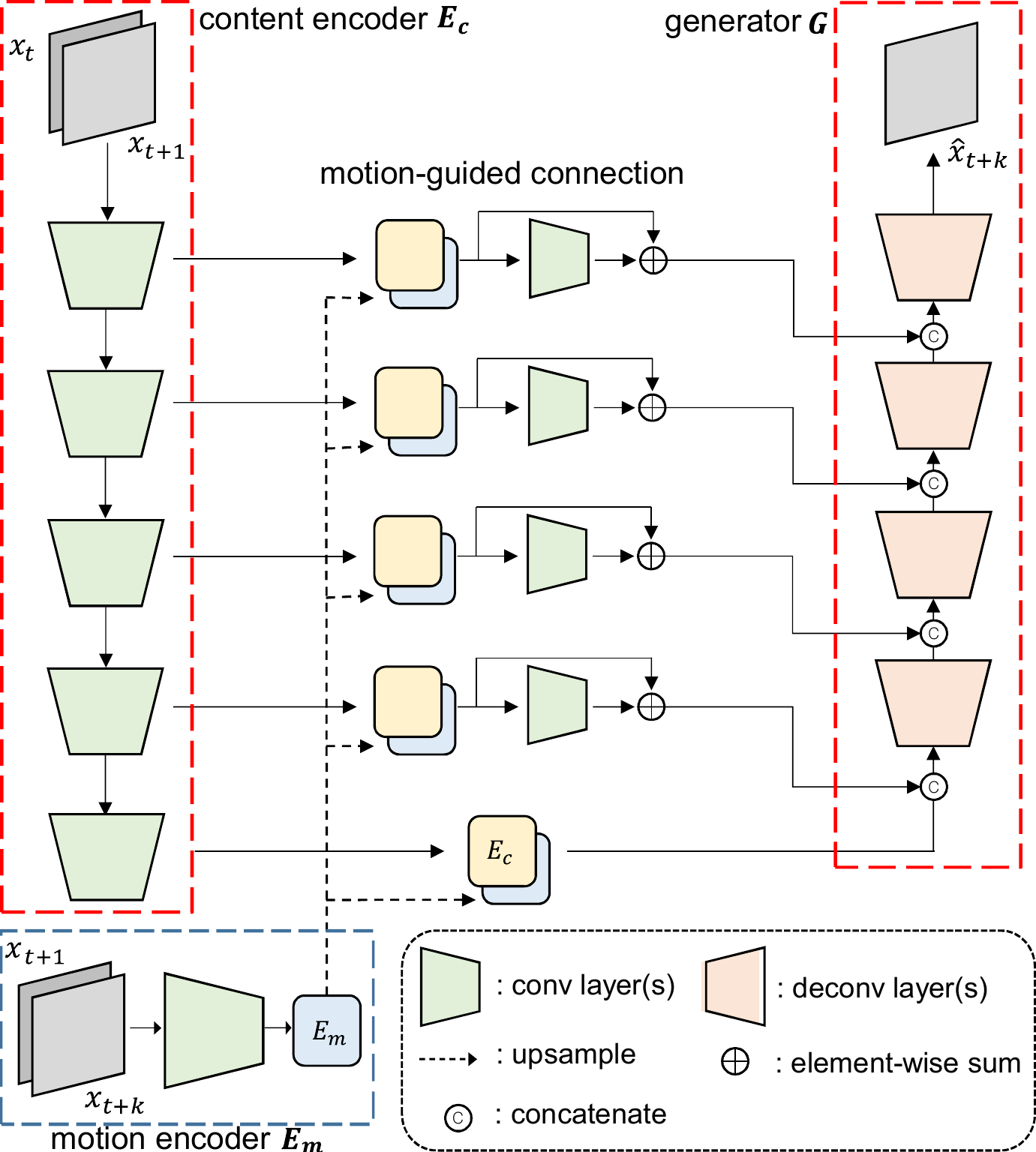}
  
  \caption{\label{figure1} Architecture of MSnet, showing multi-scale motion-guided connections.}
\end{wrapfigure}

The structure of MSnet is presented in Figure~\ref{figure1}. One encoder extracts content features and another extracts motion features. From these features, a generator reproduces $x_{t+k}$. 
Specifically, The content encoder $E_c$ obtains the content information $E_c(x_{t}, x_{t+1})$ from two successive frames $x_{t}$ and $x_{t+1}$. The motion encoder $E_m$ extracts motion information $E_m(x_{t+1}, x_{t+k})$ from frames $x_{t+1}$ and $x_{t+k}$, which do not have to be adjacent. The generator $G$ reproduces the last frame of the input $x_{t+k}$. The motivation for these settings is shown in the appendix.

\textbf{Motion-guided Connection: }The generator $G$ is connected to the content encoder $E_c$ by block-wise motion-guided connections, which play a similar role to the skip connections in UNet~\cite{ronneberger2015u}, but each motion-guided connection performs an additional convolution operation guided by motion feature. This reduces ghosting in the reproduced frame: a standard skip connection tends to preserve information about previous frames $x_{t}$ and $x_{t+1}$ (but not $x_{t+k}$), which causes a ghost of $x_{t}$ and $x_{t+1}$ to remain in the reproduced $x_{t+k}$. We concatenate the features of each convolutional block and a bi-linearly upscaled motion feature $E_m(x_{t+1}, x_{t+k})$, and then pass the concatenated features into a $1\times1$ convolutional layer to adjust the number of channels, and add residual connections. These motion-guided connections use motion information to modify the spatial information from previous frames, so that it can be effectively transferred to the target frames.

\textbf{Discriminators for adversarial training: }We apply adversarial learning to train MSnet, with three discriminators. For realistic and sharp results, we use a frame discriminator. We use two additional discriminators to disentangle motion and content features. More details of these are given in Section~\ref{trainingprocedure}.

\subsubsection{Training Procedure}\label{trainingprocedure}

Based on the intuitive assertions presented in the Introduction, we define the following objective terms: To train the encoders, we use
\vspace{-3pt}
\begin{align}\label{equation9}
   L_1=L_{\text{rec}} +\alpha L_{\text{rev}} + \beta (L_{\text{advC}} + L_{\text{advM}} + L_{\text{advF}}),
\end{align}
where $\alpha$ and $\beta$ are hyperparameters. To train the discriminators, we use
\begin{align}\label{equation10}
   L_2 = L_{\text{DF}} + L_{\text{DC}} + L_{\text{DM}}.
\end{align}
We optimize $L_1$ and $L_2$ alternately. The loss terms in $L_1$ and $L_2$ are described below. In what follows, $\hat{x}_{t+k}$ denotes the reproduced frame $G(E_c(x_t, x_{t+1}), E_m(x_{t+1}, x_{t+k}))$.

\textbf{Reconstruction and time-reversal Losses: }
Based on the `Reproducibility' and `Time-reversibility' assertions presented in the Introduction, we define $L_{\text{rec}}$ and $L_{\text{rev}}$ as follows:
\begin{equation}\label{equation1}
   L_{\text{rec}} = \|\hat{x}_{t+k}- x_{t+k}\|_{2}^{2},
\end{equation}
\vspace{-7pt}
\begin{equation}\label{equation2}
   L_{\text{rev}} = \|E_c(x_t, x_{t+1}) - E_c(x_{t+1}, x_t)\|_{2}^{2},
\end{equation}
where $k$ represents the temporal distance between the target frame and the reference frame.

\textbf{Frame adversarial loss: }
DRnet \cite{denton2017unsupervised} uses mean squared error loss alone, which tends to produce blurry results in image reproduction \cite{mathieu2015deep}. We thus introduce an extra frame adversarial loss, using a technique similar to that employed in the pix2pix network \cite{isola2017image}.
The frame discriminator $D_f$ is trained to determine whether its input is a real pair of frames or not, and $D_f$ is trained by $L_{\text{DF}}$ which is expressed as follows: 
\begin{align}\label{equation3}
   L_{\text{DF}} = - \log D_f (x_t, x_{t+k}) - \log(1-D_f (x_t, \hat{x}_{t+k}))
\end{align}
The adversarial loss $L_{\text{advF}}$ expresses the extent to which synthetic frames produced by the generator $G$ manage to deceive the discriminator. The generator $G$ is trained by $L_{\text{advF}}$ to synthesize realistic frames with the aim of deceiving the frame discriminator, and $L_{\text{advF}}$ is expressed as follows:
\begin{equation}\label{equation4}
   L_{\text{advF}} = -\log D_f (x_t, \hat{x}_{t+k}).
\end{equation}

\textbf{Disentangling adversarial loss: }
The notion of `Separability of features' described in the Introduction is realized by the content discriminator $D_c$ and motion discriminator $D_m$.
The content discriminator is trained to determine whether two motion features come from the same video, which requires it to discover the content information in these features.
Thus, to deceive the content discriminator, the motion encoder must generate motion features that contain as little content information as possible.
We train the content discriminator to discover content information in motion features using the loss $L_{\text{DC}}$, and the loss $L_{\text{advC}}$ is used to train the motion encoder in such a way that the motion discriminator cannot make a decision, which means that the entropy becomes maximized. Note that Eq~\ref{equation6} can be simplified to $L_{\text{advC}} =-\log x-\log (1-x)$ if we set $x=Dc(…)$, and the function has the minimum value when $x=1/2$. 
The result is that the motion encoder obtains a pure motion feature. These two losses are formulated as follows:
\begin{figure}[t]
    \centering
    \includegraphics[width=0.88\linewidth]{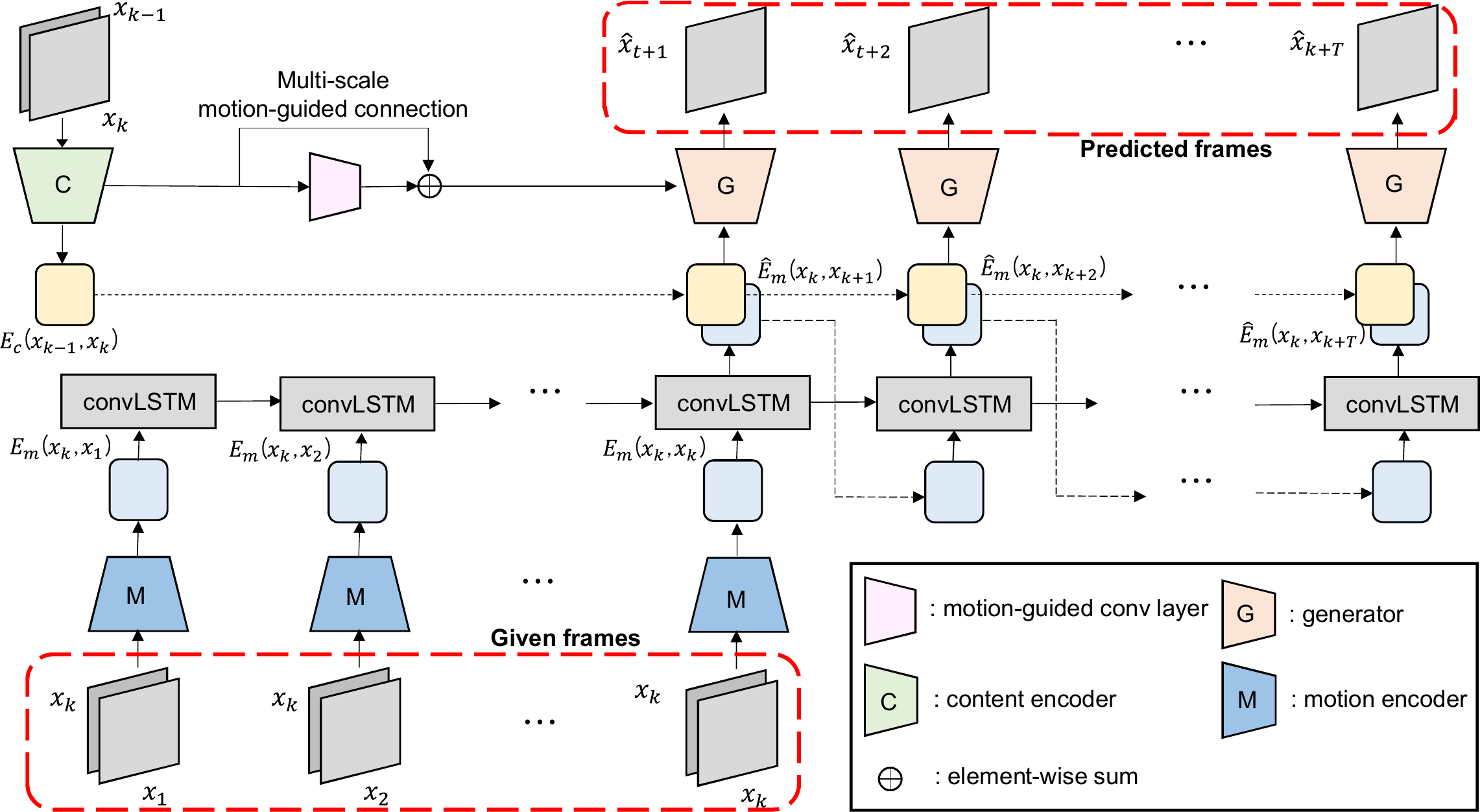}
    \caption{The frame prediction network. Given $k$ frames, the network predicts upcoming $T$ frames.} 
\vspace*{-10pt}
\label{fig_2}
\end{figure}
\small{
\begin{align}\label{equation5}
   L_{\text{DC}} =& - \log D_c (E_m(x_a, x_{a+1}), E_m(x_b, x_{b+1})) -\log(1-D_c (E_m(x_a, x_{a+1}), E_m(y_b, y_{b+1})))
\end{align}
\begin{align}\label{equation6}
	L_{\text{advC}} =& - \log D_c (E_m(x_a, x_{a+1}), E_m(x_b, x_{b+1})) - \log (1 - D_c (E_m(x_a, x_{a+1}), E_m(x_b, x_{b+1})))
\end{align}}
where $a$ and $b$ are different frame numbers, and $x$ and $y$ are different videos.


In a similar way, the motion discriminator is trained to determine whether two content features are from sequential or non-sequential frames, which requires it to discover the motion information from the content feature.
The content encoder can deceive the motion discriminator if it generates content features that do not contain motion information. 
We train the motion discriminator to discover motion information in the content feature by the loss $L_{\text{DM}}$, and the content encoder is trained to deceive the motion discriminator by the loss $L_{\text{advM}}$; so that the content encoder can obtain a pure content feature. These two losses are formulated as follows:
\begin{align}\label{equation7}
   L_{\text{DM}} =&	 - \log D_m (E_c(x_a, x_{a+1})) - \log (1 - D_m(E_c(x_a, x_b))),
\end{align}
\begin{align}\label{equation8}
	L_{\text{advM}} =& - \log D_m (E_c(x_a, x_{a+1})) - \log (1 - D_m(E_c(x_a, x_{a+1}))),
\end{align}
where $x_a$ and $x_{a+1}$ are sequential frames, and $x_a$ and $x_b$ are non-sequential frames.

\subsection{Video Frame Prediction}
\label{futureframepred}
We apply the motion and content encoders trained during frame reproduction to video prediction. MSnet is given $k$ frames $(x_1, \cdot\cdot\cdot, x_k)$ and trained to predict the following $T$ frames $(x_{k+1}, \cdot\cdot\cdot , x_{k+T})$, using the network illustrated in Figure~\ref{fig_2}. The motion encoder extracts motion features from the pairs $(x_k, x_1), (x_k, x_2), \cdot\cdot\cdot$, $(x_k, x_k)$, and the content encoder extracts content features from $(x_{k-1}, x_k)$. Note that the first frame in each pair is always $x_k$ during motion extraction. A convolutional LSTM network (cLSTM) \cite{xingjian2015convolutional} takes the motion features $E_m(x_k, x_t) (1 \leq t\leq k)$ extracted from each given pairs of frames and predicts the motion features of the subsequent frames $\hat{E}_m(x_k, x_{t+1})$ until the $k^{\text{th}}$ frame. 
\begin{equation}\label{equation11}
   \text{cLSTM}(E_m(x_k, x_t)) = \hat{E}_m(x_k, x_{t+1})\;(1 \leq t \leq k).
\end{equation}
For subsequent unknown frames, the predicted motion features are fed back into the cLSTM and the motion features of the next upcoming frames are predicted. By repeating this step, we can predict the motion features of the following $T$ frames.
\begin{equation}\label{equation12}
   \text{cLSTM}(\hat{E}_m(x_k, x_t)) = \hat{E}_m(x_k, x_{t+1})\;  (k < t < T)
\end{equation}
The cLSTM is trained using the following objective function:
\small{
\begin{equation}\label{equation13}
   L_{\text{lstm}} = \|\text{cLSTM}(E_m(x_k, x_k)) - E_m(x_k, x_{k+1})\|^2  +\sum_{t=k+1}^{T-1}\|\text{cLSTM}(\hat{E}_m(x_k, x_t)) - E_m(x_k, x_{t+1})\|^2
\end{equation}}
Finally, the generator produces $\hat{x}_t$ from the $t^{\text{th}} ~(t>k)$ predicted motion features $\hat{E}_m(x_k, x_t)$, together with the content features $E_c(x_{k-1}, x_k)$. By repeating this step, we can generate the required number of upcoming frames.

\vspace{-10pt}
\section{Experiments}
\begin{figure*}[t]
  \centering
  \includegraphics[width=0.99\linewidth]{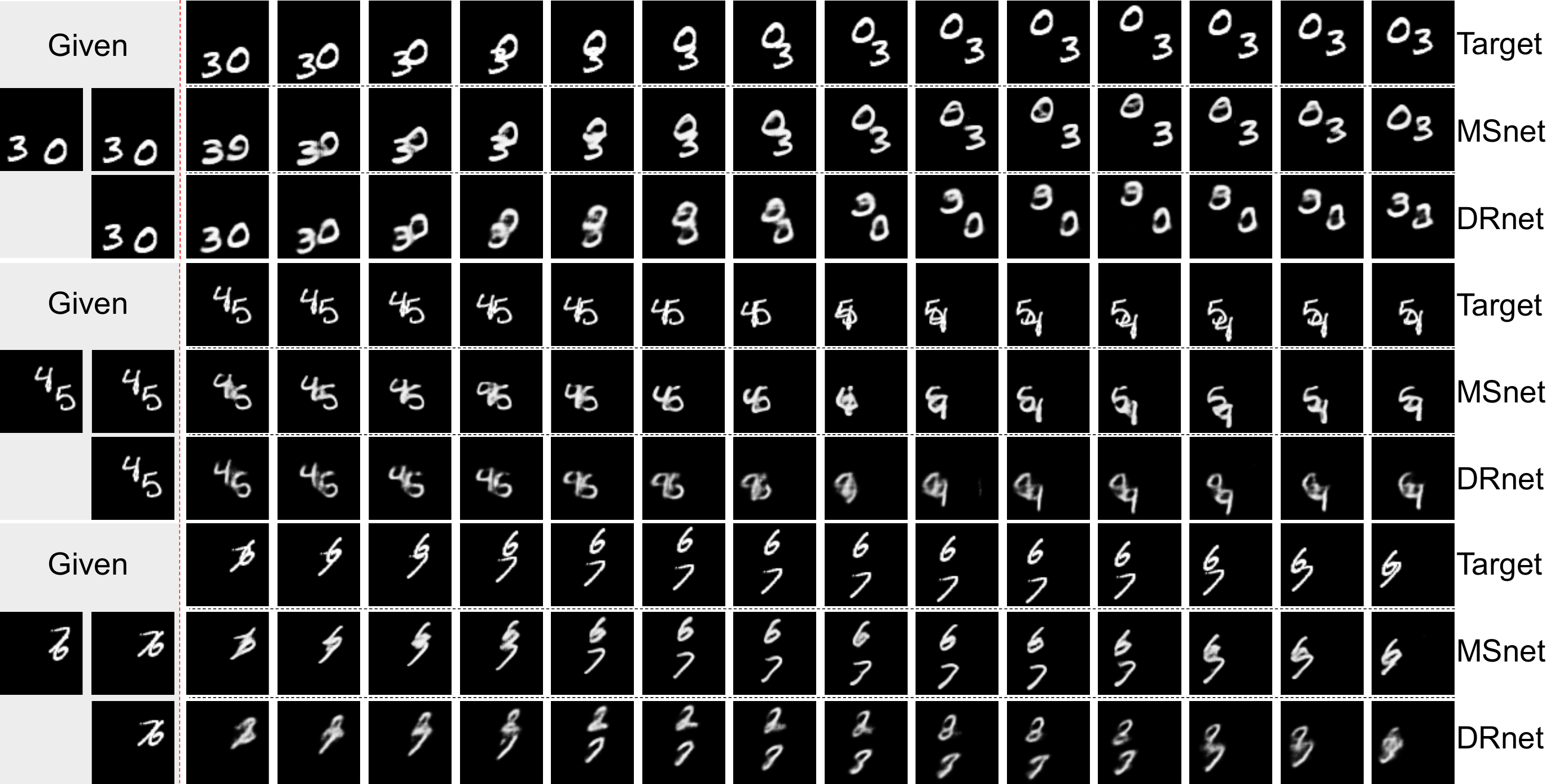}
  \vspace{-8pt}
  \caption{Qualitative results of frame reproduction task on the Moving MNIST dataset.}
  \vspace{-12pt}
  \label{fig_4}
\end{figure*}
We performed experiments using the Moving MNIST and KTH datasets~\cite{srivastava2015unsupervised,schuldt2004recognizing}. 
First, we performed frame reproduction using the Moving MNIST to compare MSnet with DRnet~\cite{denton2017unsupervised}. Then, we present frame reproduction, frame prediction and disentangling experiments (feature-based nearest retrieval and t-SNE visualization) on the KTH dataset. 

\begin{figure}[t]
  \centering
  \includegraphics[width=1.0\linewidth]{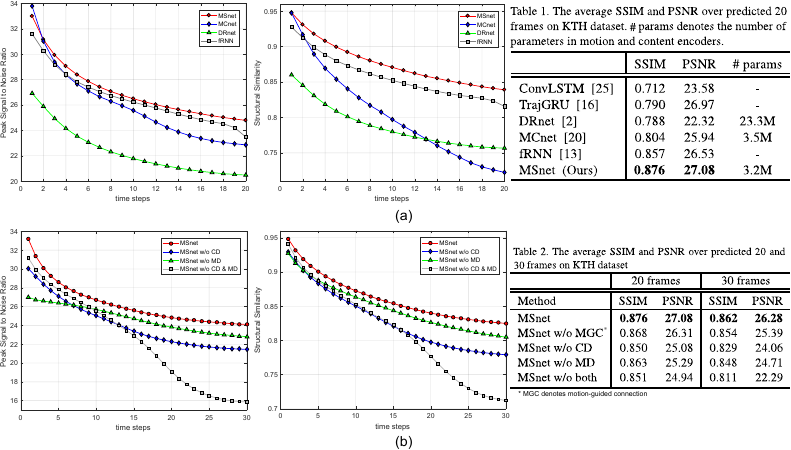}
    \vspace{-20pt}
    \caption{\label{fig_6} Quantitative results of the frame prediction on the KTH dataset. (a) Comparison with state-of-the-art methods. (b) Comparison with ablation settings.}    \vspace{-8pt}
\end{figure}
\vspace*{-8pt}
\subsection{Moving MNIST}
The Moving MNIST dataset~\cite{srivastava2015unsupervised} contains 10,000 video sequences, each consisting of 20 frames. In each video sequence, two digits move independently around the frame, which has a spatial rsolution of $64 \times 64$ pixels. The digits frequently intersect with each other and bounce off the edges of the frame. We used 8,000 sequences for training and 2,000 for testing. We used motion features with a $4 \times 4$ spatial map and $4$ channels, and content features with a $4 \times 4$ spatial map and $8$ channels. We use more channels for the content features because the motions occurring in the Moving MNIST videos are not as complicated as those in natural videos. We used values of the temporal distance $k$ between $0$ and $5$ in the frame reproduction process, and we set $\alpha=1.0$ and $\beta=3.3 \times 10^{-5}$ in Eq.~\eqref{equation9}.

Qualitative results from this experiment are shown in Figure~\ref{fig_4}.  Note that Denton and Birodkar (2017) used self-generated colored digits to train DRnet, thus we re-trained it with the publicly available Moving MNIST data to make a fair comparison with MSnet. MSnet obtains content features from the given frames and motion features from the last given and target frames. DRnet obtains content features from the given frames and pose features from the target frames. 

In the first example of Figure~\ref{fig_4}, DRnet generates digits in the wrong places. We attribute this to the way in which DRnet encodes temporal attributes into pose features, and not into motion features, like those used by MSnet. This suggests that motion is a more natural attribute of video than pose. In the second example, DRnet produces blurry results where the two digits overlap in target frames. In the third example, DRnet cannot identify two digits which overlap in a given frame.
MSnet can identify these overlapping digits correctly because it obtains content features from two frames. More results with Moving MNIST are presented in the supplementary material.

\subsection{KTH Dataset}
The KTH dataset~\cite{schuldt2004recognizing} contains videos of 25 people performing six actions. For our experiments, we resized the frames in the videos to $128 \times 128$ pixels. We used person 1-16 for training and person 17-25 for testing, following the widely used baseline method MCnet~\cite{villegas2017decomposing}. We used SSIM, PSNR, and inception score as evaluation metrics. We used motion features with a $8\times 8$ spatial map and $8$ channels and content features with a $8\times 8$ spatial map and $8$ channels. We used values of the temporal distance $k$ between $0$ and $10$ in the frame reproduction process. We set $\alpha=1.0$ and $\beta=4 \times 10^{-5}$ in Eq.~\eqref{equation9}. For the following figures, we denote the motion and content discriminators as MD and CD, respectively.
\subsubsection{Video frame prediction}
We used the same experimental settings used in MCnet for frame prediction experiments. All baseline networks were trained by taking 10 frames from the KTH dataset and using them to predict the following 10 frames. During testing, 3,559 sequences of $30$ frames were used: $10$ given frames and $20$ frames to be predicted. The published DRnet model was trained on person 1-20, so we re-trained DRnet on person 1-16 for fair comparison with other baseline methods. Quantitative results are presented in Figure~\ref{fig_6}(a) and (b). 
\begin{figure}[t]
  \centering
  \includegraphics[width=1.0\linewidth]{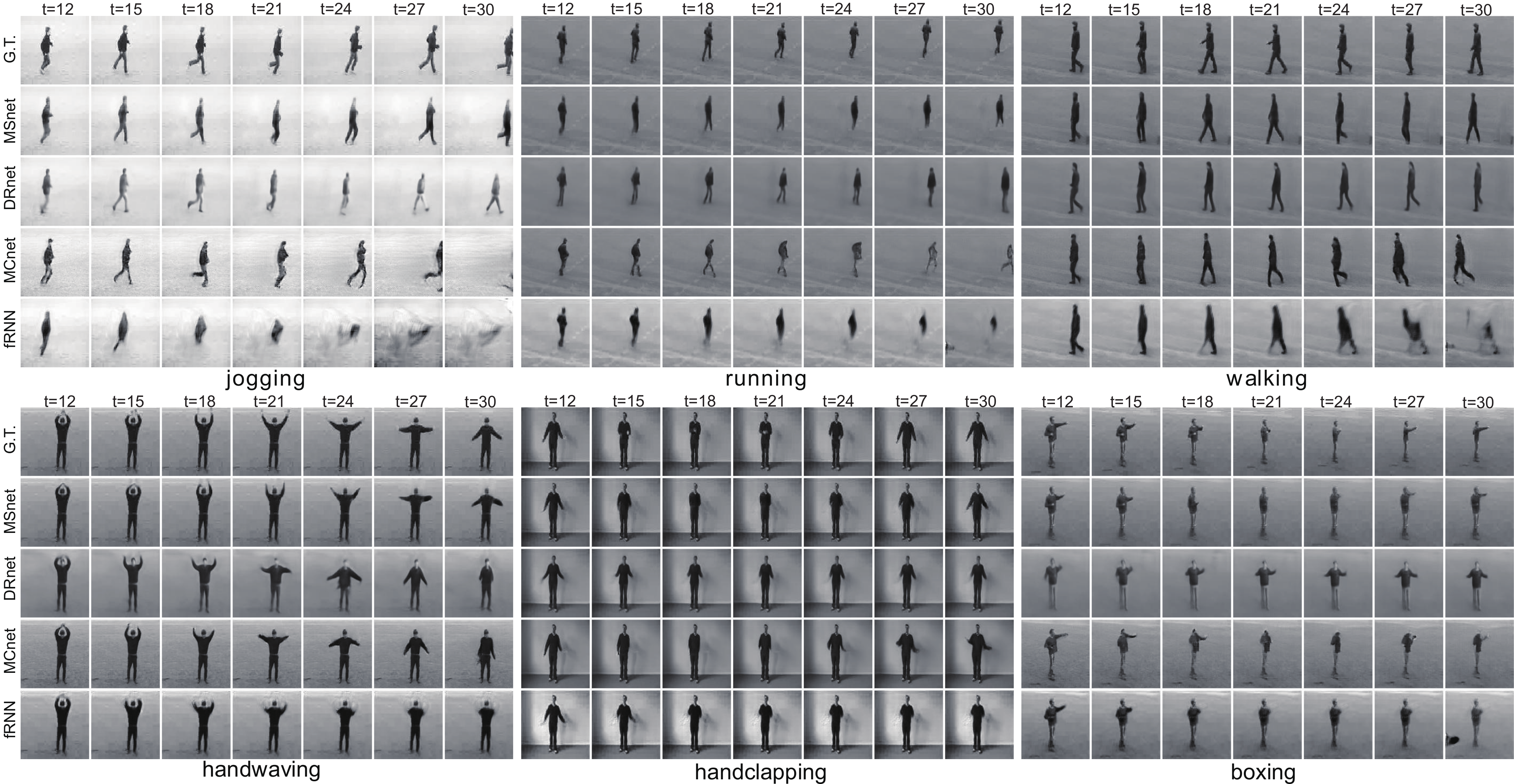}
  \vspace{-15pt}
    \caption{\label{kthsample} Qualitative results of frame prediction on the KTH dataset. Given $10$ frames, the following $20$ frames are predicted. We show every 3 frames.}
    \vspace{-7pt}

\end{figure}

Figure~\ref{fig_6}(a) shows that MSnet obtained better results than other state-of-the-art methods on three evaluation metrics even though it has simpler motion and content encoders. Note that we do not report the number of parameters for ConvLSTM~\cite{xingjian2015convolutional}, TrajGRU~\cite{shi2017deep}, and fRNN~\cite{oliu2017folded}, because it does not decompose videos into motion and content streams. Qualitative results are shown in Figure~\ref{kthsample}. DRnet produces the wrong motion in the boxing video, and changes the identity of the person in the handwaving video. We attribute these problems to DRnet's use of a basic UNet and one-directional suppression. MCnet produces a person with an unrealistic shape, which we attribute to its poor disentangling of features and a lack of semantic information in its motion features. fRNN has difficulty when the person in the frame makes a large motion, and we attribute this to the way in which it considers motion and content information simultaneously. These results suggest that meaningful features are obtained by mutual suppression and motion-guided connection. More results are presented in the supplementary material.

In ablation experiments, we removed each discriminator in turn to show the effects of mutual suppression on the disentangled features. Figure~\ref{fig_6}(b) shows that the results from MSnet are worse when either the motion or content discriminator is removed. Without the content discriminator (blue and gray lines), prediction performance drops significantly across subsequent frames because the motion encoder generates impoverished motion feature. With only the content discriminator (green line), the content encoder cannot extract meaningful content features, so it performs poorly on the first predicted frame. However, its performance does not drop significantly across subsequent frames as meaningful motion features can be extracted with the content discriminator. These results demonstrate how mutual suppression disentangles motion and content features.

\subsubsection{Disentangling Experiments}
We present t-SNE visualization~\cite{maaten2008visualizing} and feature-based nearest retrieval results to show the disentangling effect. 

\textbf{Feature-based frame retrieval: }
The aim of feature-based frame retrieval is to fetch the frame which is closest to a query frame in terms of motion and content features. More formally, 
\begin{equation}\label{equation13}
   x_{\text{r}} = \argmin_x \|E(x_q) - E(x) \|\\,~~~~~E \in \{ E_c, E_m\},
\end{equation}
where $x_\text{q}$ is a query frame, $x$ is a frame other than the query frame, and $x_\text{r}$ is the retrieved frame.
\begin{figure*}[t]
  \centering
  \includegraphics[width=1.0\linewidth]{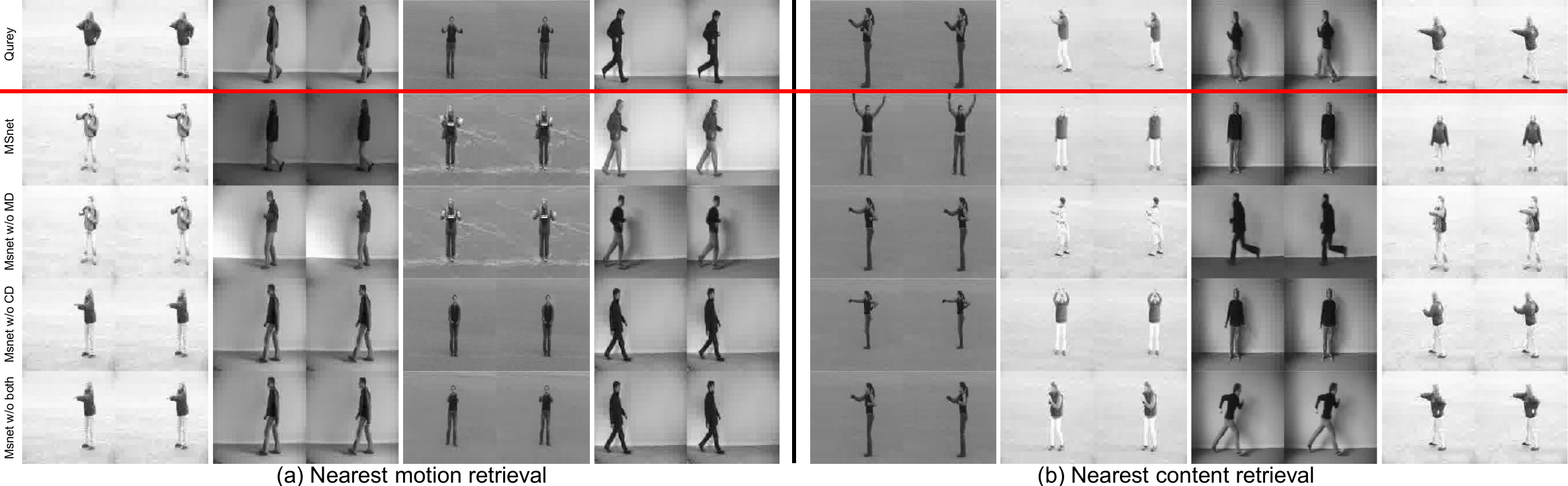}
  \vspace{-15pt}
  \caption{Feature-based nearest frame retrieval results.}
  \label{fig_nearest}
    \vspace{-10pt}

\end{figure*}
 \begin{figure*}[t]
  \centering
  \includegraphics[width=1.0\linewidth]{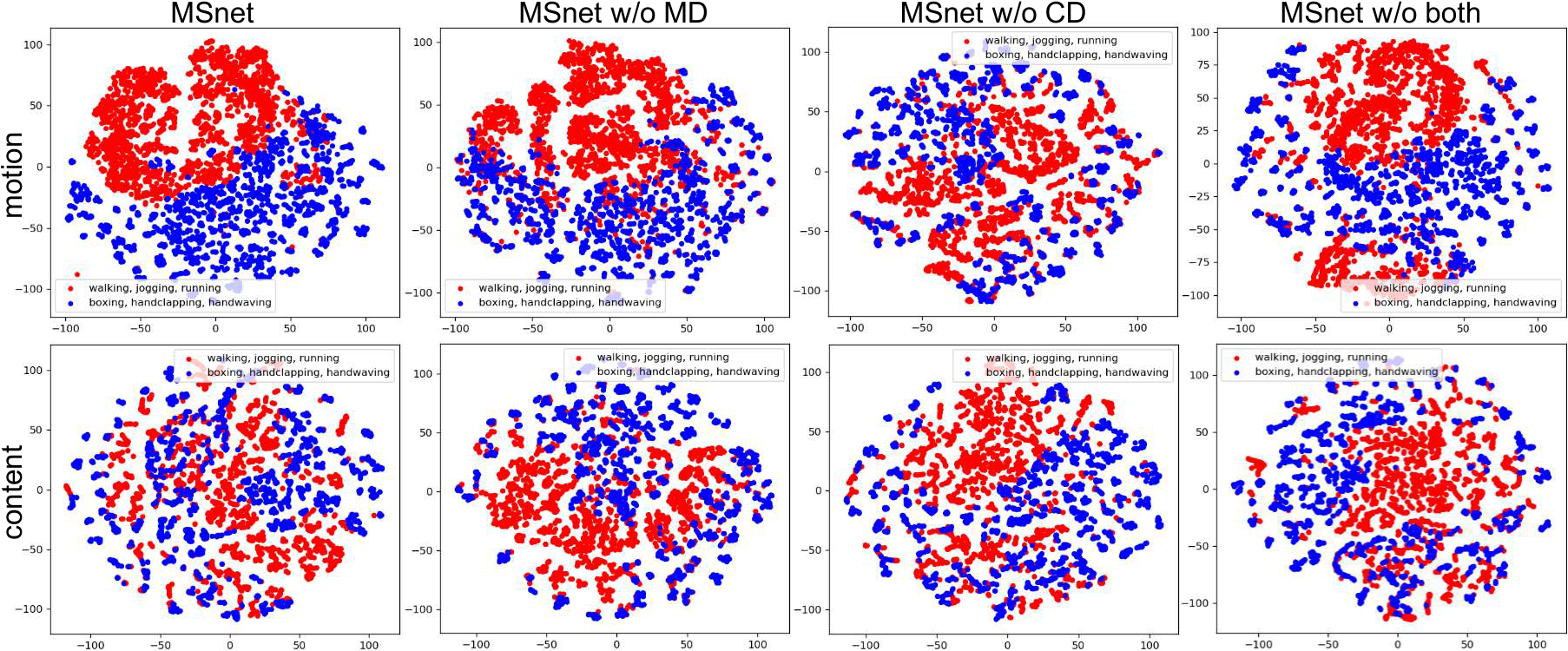}
  
  \caption{Visualization by t-SNE dimensional reduction of dynamic actions~(walking, jogging, and running) (the red points), and static actions~(boxing, handclapping, and handwaving) (the blue points), with and without the content and motion discriminators. The first row shows the distribution of motion features, and the second row shows the distribution of content features.}
   \vspace{-10pt}
  \label{fig_tSNE}
\end{figure*}

Figure~\ref{fig_nearest} shows the results of nearest motion and content retrieval. Ideally, nearest motion retrieval should retrieve the most similar motion regardless of the content (the identity of the person and the background), and content retrieval should retrieve the most similar content regardless of any motion. As shown in Figure~\ref{fig_nearest}(a), MSnets with the content discriminator (the second and third rows) retrieve the most similar motions regardless of the identity the person and the background, because the content discriminator helps the motion encoder to extract pure motion features. In Figure~\ref{fig_nearest}(b), MSnets with the motion discriminator (the second and fourth rows) retrieve the most similar content regardless of the motions, because the motion discriminator helps the content encoder to extract pure content features.

\textbf{t-SNE visualization: }As MSnet is trained in an unsupervised manner, it cannot separate similar actions, such as running and jogging. For visualization by t-SNE dimensional reduction, we thus classify walking, jogging, and running as dynamic actions, and boxing, handclapping, and handwaving as static actions. If motion and content are disentangled as intended, motion features in the same actions should be clustered, and motion features in different actions should be separated. Conversely, points corresponding to content features which do not contain motion information should not be clustered. The results of t-SNE visualization are shown in Figure~\ref{fig_tSNE}. MSnet with both discriminators produces the most clustered motion features, and the most random distribution of content features. Using the content discriminator alone, motion features are reasonably well clustered, as pure motion features can still be captured effectively. However, content features are now more clustered because the omission of the motion discriminator means that content features contain unwanted temporal information. Without the content discriminator, and without both discriminators, the results are far from what we intend.\\[-0.7em]



\vspace{-10pt}

\section{Conclusions}
We have proposed a new method for video frame prediction. We have introduced mutual suppression adversarial training to acquire disentangled motion and content representations, and applied motion-guided connection to refine the content information from previous frames for use in the prediction of upcoming frames. MSnet has been shown to obtain well-disentangled features. This lead to better results in terms of frame prediction than other state-of-the-art methods.
We have also shed light on the way in which mutual suppression disentangles features by ablation studies in the domains of t-SNE visualization and feature-based nearest frame retrieval.


\section*{Acknowledgements}
This work was supported by the National Research Foundation of Korea (NRF) grant funded by the Korea government (Ministry of Science and ICT) [2018R1A2B3001628], and the Brain Korea 21 Plus Project in 2019.

\bibliography{egbib}
\end{document}